\documentclass[runningheads]{llncs}

\usepackage{templateStyle}

\usepackage{graphicx}
\usepackage{booktabs}

\usepackage[accsupp]{axessibility}  

\usepackage{hyperref}

\usepackage{orcidlink}

\usepackage{tikz}
\usepackage{pgfplots}
\usetikzlibrary{matrix,shapes,arrows,positioning,chains}
\tikzstyle{process} = [rectangle, minimum width=3cm, minimum height=1cm, text centered, draw=black]
\tikzstyle{arrow} = [thick,->,>=stealth]
\tikzstyle{io} = [trapezium, trapezium left angle=70, trapezium right angle=110, text centered]
\usepackage{graphicx}

\usepackage{verbatim}

\begin{document}

\title{Segment Using Just One Example} 

\titlerunning{Segment Using Just One Example}

\author{Pratik Vora \and
Sudipan Saha\orcidlink{0000-0002-9440-0720}}

\authorrunning{P. Vora and S. Saha}

\institute{Yardi School of Artificial Intelligence, IIT Delhi, New Delhi, India
}

\maketitle

\begin{abstract}
Semantic segmentation is an important topic in computer vision with many relevant application in Earth observation. While supervised methods exist, the constraints of limited annotated data has encouraged development of unsupervised approaches. However, existing unsupervised methods resemble clustering and cannot be directly mapped to explicit target classes. In this paper, we deal with single shot semantic segmentation, where one example for the target class is provided, which is used to segment the target class from query/test images. Our approach exploits recently popular
Segment Anything (SAM), a promptable foundation model. We specifically design several techniques to automatically generate prompts from the only example/key image in such a way that the segmentation is successfully achieved on a stitch or concatenation of the example/key and query/test images. Proposed technique does not involve any training phase and just requires one example image to grasp the concept. Furthermore, no text-based prompt is required for the proposed method. We evaluated the proposed techniques on building and car classes.
\end{abstract}

\section{Introduction}
\label{sectionIntroduction}

Semantic segmentation is an important task in both computer vision \cite{hao2020brief} and Earth observation \cite{audebert2016semantic}. While there have been several works on supervised semantic segmentation \cite{kemker2018algorithms,diakogiannis2020resunet}, many Earth observation tasks work under constraints of limited data \cite{huang2020single} and thus unsupervised semantic segmentation \cite{saha2022unsupervised} is also a topic of interest for the Earth observation researchers. However, unsupervised semantic segmentation methods \cite{saha2022unsupervised,cho2021picie} are in essence clustering  and their obtained segmentation outputs do not  explicitly map to target class. 
Somewhat relaxing the conditions of unsupervised semantic segmentation, an example of nearly unsupervised semantic segmentation that also provides a mapping with desired target class(es) is the case when there is one training image available. As an example, imagine that we are interested to find buildings of a specific roof type. We are provided with an image that contains such roof type and its corresponding binary mask showing the pixels belonging to this roof type in one cluster and all other pixels in the other cluster. Then using this image as example or key image, we can retrieve from any query image the pixels belonging to the same roof type. Such an one-shot scenario can be useful in many applications, including in disaster management where rapid identification of specific targets can be important for planning and resource allocation. To better illustrate this situation, consider an unmanned aerial vehicle (UAV) deployed in a disaster management operation. The UAV is provided with one example image for each target category it needs to find. Ideally, the UAV should be able to use this single example to identify targets without requiring any additional training.
\par
An interesting recent development in computer vision is the development of the foundation models. Foundation models show strong generalization capability and have already found applications in semantic segmentation. While many foundation models can relate images and texts, in this work our interest is entirely on images. In more details, in the above-mentioned discussion, we only consider the case where we have example/key image and we are not interested in the cases where the key is instead expressed as text. Segment Anything (SAM) \cite{kirillov2023segment} is a vision foundation model that has shown strong capabilities for image segmentation. Taking prompts such as points, SAM can produce segmentation corresponding to the desired objects or classes. However, choosing the prompts in an autonomous manner to enhance the detection of the desired object is not a trivial process. In this work, we capitalize on the promptable segmentation capabilities of SAM to solve the above-mentioned problem of segmenting an object of interest using just one  example and without any training. As such, our work is also related to target detection \cite{qi2019ship} as we focus on one target class, however instead of producing bounding boxes, we produce segmentation masks corresponding to the target class. Such segmentation masks can be easily converted into bounding boxes, if required for a specific application.
\par

\begin{figure}[t!]
\centering
\includegraphics[height=5 cm]{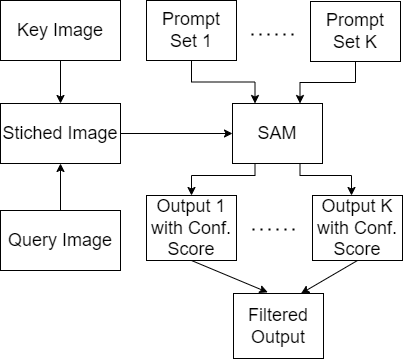}
\caption{Outline of the proposed method: given a key/example image for which the segmentation mask is known and a query image, proposed method concatenates them and feeds them to the SAM model as if they are single image. Furthermore, image-based prompts are fed to SAM that enables us to obtain the segmentation mask from the stitched image, i.e., also from the query image.}
\label{figureProposedMethod}
\end{figure}

We emphasize that our problem statement and proposed solution are tailored for scenarios where we lack prior knowledge of our targets during deployment. The flexibility of our proposed method allows us to launch specific operations with just one example image. If we already knew in advance the specific targets of interest, it would be possible to collect a dataset and train a model for that particular category. However, our approach is designed for situations where such prior information is unavailable, enabling rapid identification using minimal data. Moreover, we do not need any text-based prompts. Although text-based prompts have gained popularity recently, our focus is on scenarios where the concept is provided to a model using only a single example image, without any textual description. In many practical applications, describing the target textually might be challenging, and it could be more intuitive to simply identify the target in an image.
\par
Furthermore, our work does not rely on a training phase, making it a significant step towards reducing computational requirements and improving data efficiency. By eliminating the need for a training phase, our method not only speeds up the deployment process but also reduces the computational burden, making it suitable for real-time applications in dynamic environments.
\par
The key contributions of this work are as follows:
\begin{enumerate}
\item We present an interesting and useful problem statement where only one example image of a specific category is provided, which is to be used for semantic segmentation on any query image without any training.
\item To solve the above-mentioned problem statement, we propose a solution exploiting recently popular foundation model SAM. Furthermore, we construct several novel automatic prompt engineering techniques employing SAM.
\item We evaluate the proposed method in the context of Earth observation, specifically in context of building and car segmentation. However, we emphasize that these two classes are chosen merely for the challenges associated in delineating them, e.g., intra-class variability and different sizes. Otherwise, the practical utility of the proposed problem statement lies in its ability to handle unforeseen classes that may arise in various applications, particularly in disaster management scenarios where it is impossible to anticipate all potential targets in advance. 
\end{enumerate}
\par
We organize the rest of the paper as follows. A few related works, such as semantic segmentation and foundation models, are briefly discussed in Section \ref{sectionRelatedWorks}. Following this, Section \ref{sectionProposedMethod} discusses the proposed method for target-specific semantic segmentation with single example image. Results are discussed in Section \ref{sectionExperimentalResult}. Finally, the chapter is concluded in Section \ref{sectionConclusion}.

\section{Related works}
\label{sectionRelatedWorks}
\subsection{Supervised semantic segmentation} Widely used segmentation architectures in deep learning include fully convolutional networks (FCNs) \cite{long2015fully}, U-Net \cite{ronneberger2015u}, SegNet \cite{badrinarayanan2017segnet}, and as DeepLab \cite{chen2017rethinking}. In the domain of Earth observation images, several supervised segmentation algorithms have been introduced leveraging these architectures \cite{volpi2017dense,maggiori2017high,su2022semantic,ma2021semantic}. However, these supervised methodologies require substantial volumes of training data for supervised learning.

\subsection{Semantic segmentation with limited labels} Several works in the literature attempt to train semantic segmentation models with partial or no label, however generally with lower performance than supervised models. The work in \cite{hua2021semantic} proposes a method to train semantic segmentation models with sparse annotations. An unsupervised semantic segmentation model is proposed in \cite{saha2022unsupervised} exploiting contrastive learning.  A Siamese network based method for one shot semantic segmentation in computer vision is proposed in \cite{zhao2021one}.

\subsection{Foundation models} The vision foundation models represent an important advancement in computer vision, reshaping how we analyze visual data \cite{awais2023foundational}. These models, which primarily rely on self-supervised learning on vast datasets, demonstrate robust generalization capabilities across a wide range of applications. Vision Transformers \cite{dosovitskiy2020image} are commonly utilized in these models for their effectiveness in capturing long-range dependencies within images \cite{wang2023internimage}.
Among other foundation models, SAM \cite{kirillov2023segment} is designed for promptable semantic segmentation and thus directly related to and used in our work. There are already some works using SAM for Earth observation tasks \cite{chen2024rsprompter,gui2024evaluating}, however generally used in supervised context unlike our work. The work in \cite{chen2024rsprompter} trains category-specific prompt generator that enables generation of prompts to be fed to SAM. The work in \cite{gui2024evaluating} uses SAM for identifying green spaces, however by using dataset augmentation and fine tuning. Contrary to the existing works, our method is unsupervised and just uses a single example, which is not used for training. Furthermore, our method does not use any category-specific training unlike \cite{gui2024evaluating}.

\section{Proposed Method}
\label{sectionProposedMethod}

\subsection{Problem}
Let us assume that we have an example/key image-label pair $X_{key},Y_{key}$. The label image $Y_{key}$ is a binary image that consists of two classes: all pixels belonging to a target category and all other pixels that do not belong to this category. As such this, example/key image can also be called training image, however we do not prefer to call so as there is no training step involved in our proposed method. Using this single example, we want to segment any query/test image $X_{test}$. We assume that the key and test images have same dimension, though this assumption can be easily relaxed.
\par
To address the above-mentioned problem, we use SAM (Section \ref{sectionSAM}), to which we feed a stitched image comprising of key and query images (Section \ref{sectionImageStitching}) and prompts designed in one of several ways (Section \ref{sectionSAMPrompts}). By repeating this process $K$ times, we form an ensemble (Section \ref{sectionEnsemble}) and further aggregate the ensembles (Section \ref{sectionEnsembleAggregation}) and apply post-processing (Section \ref{sectionPostProcessing}). The outline of the proposed idea is demonstrated in Figure \ref{figureProposedMethod}.

\subsection{SAM}
\label{sectionSAM}
Our method uses SAM which comprises of the following components:
\begin{itemize}
    \item \textbf{Image Encoder}: Given an input image, the image encoder produces its embedding that captures the visual content of the image.
    \item \textbf{Prompt Encoder}: The prompt encoder encodes the prompt input to the model. There can be several types of prompts, however in this work we only consider point and mask  prompts. 
    \item \textbf{Mask Decoder}: The mask decoder takes the representations generated by the encoders and produces the segmentation mask.
\end{itemize}

\begin{figure*}[t!]
\centering
\includegraphics[height=2.6 cm]{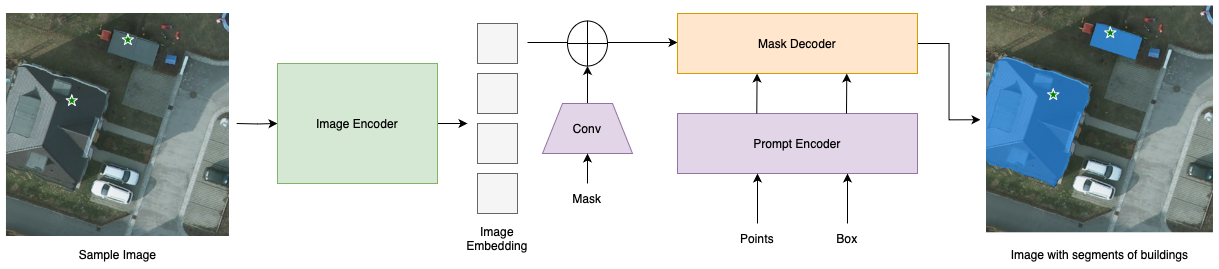}
\caption{SAM pipeline (green: positive prompt, red: negative prompt). The image encoder obtains representation of the image whereas prompt encoder obtains the representation of the prompt inputs. Using this information, the decoder obtains the segmented image.}
\label{figureSAMPipeline}
\end{figure*}

SAM generally ingests a single image and the prompts. We must emphasize that the prompts can be both positive and negative, corresponding to the category of interest and to the other categories, respectively. SAM produces the desired segmentation mask and a confidence score.  The pipeline of SAM is shown in Figure \ref{figureSAMPipeline}.

\subsection{Image concatenation}
\label{sectionImageStitching}
Since we are working with two images: $X_{key}$ and $X_{test}$, we stitch them together, i.e., merely concatenate them side by side to produce a new image $X_{cat}$ which we then feed to SAM. Recall that the segmentation mask for the $X_{key}$ is known. Thus, if we assume $Y_{cat}$ is the segmentation mask for $X_{cat}$, then we already know half of $Y_{cat}$ (which is same as $Y_{key}$) and our task is to estimate the other half ($Y_{test}$).

\begin{figure}[t!]
\centering
\begin{subfigure}{0.7\linewidth}
\centering
         \fbox{\includegraphics[height=2 cm,width = 5cm]{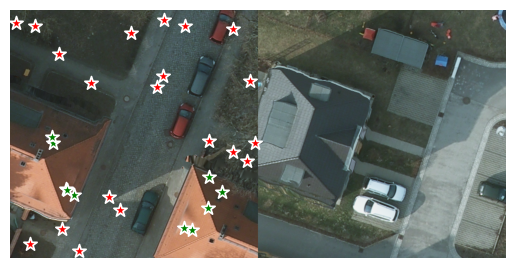}}
            \caption{}
            \label{figureKeyOnlyPrompts}
\end{subfigure}

\begin{subfigure}{0.7\linewidth}
\centering
         \fbox{\includegraphics[height=2 cm,width = 5cm]{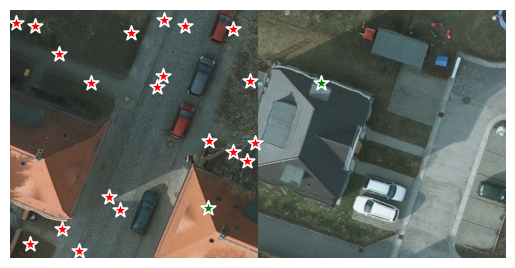}}
            \caption{}
            \label{figureKeyPromptsAndAdditionalPositivePromptFromTest}
\end{subfigure}

\begin{subfigure}{0.7\linewidth}
\centering
         \fbox{\includegraphics[height=2 cm,width = 5cm]{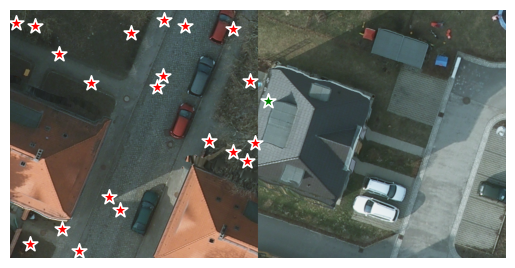}}
            \caption{}
            \label{figureNegativeKeyPositiveTestPrompt}
\end{subfigure}

\begin{subfigure}{0.7\linewidth}
\centering
         \fbox{\includegraphics[height=2 cm,width = 5cm]{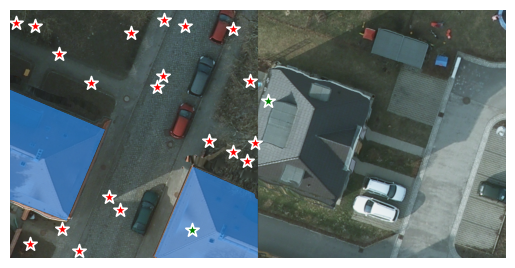}}
            \caption{}
            \label{figureMaskedKeyPositiveTestPrompt}
\end{subfigure}

\caption{Proposed prompt techniques: (a) key only prompts, (b) key prompts and positive prompts from query/test , (c) negative prompts from key and positive prompts from query/test and (d) masked key prompts and positive prompts from query/test. Left image is the key image and right image is the query/test image. Positive prompts are shown in green and the negative prompts are shown in red.}
\label{figureDifferentPromptTechniques}
\end{figure}

\subsection{Designing prompts for SAM}
\label{sectionSAMPrompts}
We postulate that the above-mentioned task can be handled by cleverly choosing the prompts from the known half of  $Y_{cat}$ and exploiting SAM to produce the segmentation mask for the unknown half. Towards this, we design four different prompt techniques that are all image-based and requires no textual input. The first technique incorporates point-based prompts from the key image only. The following two techniques also cleverly incorporates the point-based prompts from the query image. The fourth technique additionally uses mask prompt from the key image.

\subsubsection{Key-only prompts}
As discussed above, one-half of $Y_{cat}$ (i.e., $Y_{key}$) is known to us and we can simply choose positive and negative point prompts from $Y_{key}$, by randomly sampling a few points that that belong to the category of interest (positive prompts) and choosing some other points that do not belong to the category of interest (negative prompts).  This is the most natural choice and positive and negative prompts are all correct. However, one issue with such choice of prompt is that all positive (and also negative) prompts are chosen from one half of the stitched input. This prompt technique is demonstrated in Figure \ref{figureKeyOnlyPrompts}. Though intuitive, in such an input scenario, SAM may get spatially biased and obtain an impression that the category of interest is likely to be present in only in this half of the stitched input. We will later see in the Section \ref{sectionExperimentalResult} that this is indeed a critical issue. 

\subsubsection{Key prompts and additional positive prompt from query/test}
\label{subSectionPrompt2}
As discussed above, the key limitation of the key-only prompt engineering is that SAM may get a spatial bias towards only the key image. To mitigate this issue, we need to somehow pursue SAM to also start considering the test image. This can be done by keeping the prompt engineering similar to above and additionally choosing positive point prompt from the test image. However, we do not know the segmentation mask of the test image and hence the question remains that how do we choose positive prompts from the test image. As there is no way to ensure this, we will simply choose some point randomly sampled from query/test scene as positive prompt. Note that now this positive prompt may or may not be correct. In other words, some points chosen as positive prompt from the query/test image may turn out to be actually not belonging to the category of interest and hence actually a negative point erroneously fed as positive prompt. This might discourage us to design such prompt engineering, however we will see in Section \ref{sectionEnsemble} that this challenge can be tackled in a systematic manner. This prompt technique is shown in Figure \ref{figureKeyPromptsAndAdditionalPositivePromptFromTest}.

\subsubsection{Negative prompts from key and positive prompt from test}
In this approach, we go a step further and we choose positive prompt from the query/test (like above) and only negative prompts from the key, i.e., we do not choose any positive prompts from key. Since our negative prompts are sampled from key, they are all correct. Since our positive prompt is sampled from query/test for which we do not know the segmentation mask yet, it can be incorrect. However, this issue can be handled in a systematic manner, as described in Section \ref{sectionEnsemble}. This prompt technique is shown in Figure \ref{figureNegativeKeyPositiveTestPrompt}.

\subsubsection{Masked key prompts and positive prompt from query/test}
In this approach we accompanied the prompt from the second approach with an input mask of the key. Let's say that we are interested in finding segment for buildings, so we will have the ground truth mask of all the buildings for the key image. This key mask will be a binary matrix of size equal to that of key image and will have value 1 where there is a building in the key image and 0 everywhere else. For the query image, we do not have any ground truth mask and so we will take a zero matrix of size equal to the query image. These two masks were stitched together to form a combined mask prompt. An example of this prompt technique is shown in figure \ref{figureMaskedKeyPositiveTestPrompt}.

\subsection{Using ensemble and confidence score}
\label{sectionEnsemble}
As discussed above, all the three prompt engineering techniques have some merits and demerits. The first one selects all correct positive and negative prompts, however has a spatial bias towards the key image. The other two reduces the spatial bias, however may choose incorrect positive prompts. Thus, feeding $X_{cat}$ and the chosen prompts using any of the above techniques to SAM does not guarantee generation of desired $Y_{cat}$. Here we adopt an ensemble technique. Sampling some positive and negative points and feeding the inputs once to SAM can be considered as one run of SAM. Similarly, we can sample another set of positive and negative points and feed to SAM again. By repeating this process $K$ times, we can obtain $Y^1_{cat},...,Y^K_{cat}$. In addition to the segmentation output, SAM also provides a confidence score. Recall the discussion that in our second (and third) prompt engineering techniques, we are not certain about correctness of the positive prompts, which might lead to incorrect segmentation result. However, since our negative prompts are certainly correct, if randomly chosen positive prompts disagrees in concept with them, then the produced segmentation mask generally has a low confidence score. Thus, running the model $K$ times, with different randomly sampled prompts, we may expect that some of these $K$ runs will have relatively correctly chosen prompts, which will be aggregated in the next step to obtain the test image segmentation mask: $Y_{test}$.

\subsection{Aggregation}
\label{sectionEnsembleAggregation}
After receiving the results for $K$ number of prompts, we need to devise a strategy to aggregate them to form a single result. For this, we propose two different strategies: best selection and Confidence Weighted Majority Voting (CWMV).
\subsubsection{Best selection}
In this approach, we simply choose the segmentation result with highest confidence score. This method operates on the assumption that the output with the highest confidence score is superior to all other possible outputs. Essentially, the confidence score represents a measure of reliability in the result. By selecting the outcome with the highest score, the method assumes that this particular outcome is the most accurate or optimal among all the available results.

\subsubsection{CWMV}
This strategy involves using confidence weighted average of all the results. This does not carry the assumption of the previous strategy that the output with highest confidence score is superior to all other possible outputs. CWMV \cite{meyen2021group} is an ensemble learning technique used in machine learning to improve the accuracy of predictions by leveraging the confidence levels of individual classifiers.

For a class $C$, let $\{Y^i_{test}\}_{i=1}^{K}$ be the predicted segments for the $K$ different prompts and $\{c^i_{test}\}_{i=1}^{K}$ be their confidence scores. The weighted segment $\Tilde{Y}$ is calculated as:

\begin{equation}
    \Tilde{Y} = \sum_{i=1}^{K}{c^i_{test} * Y^i_{test}} 
\end{equation}
Now, we will have $\Tilde{Y}$ where every value $\Tilde{Y}_{ij}$ will be the aggregated score over all the prompts for (i,j) pixel in the image. We  calculate a threshold $\tau$ as:

\begin{equation}
    \tau = \frac{\sum_{i=1}^{K}{c^i_{test}}}{m}
\end{equation}

Here, m is a hyperparameter which can be set as per this criterion: if we want to eliminate x\% of the pixels with least confidence, then we can set m to 100/x. Thus, to remove 25\%, we can set m to 4. Lower the value of m, higher will be $\tau$ (the threshold) and hence more number of pixels identified as our object of interest will be removed.
For every pixel $(i,j)$ of the weighted segment $\Tilde{Y}$, the segment assignment $\Tilde{S}_{ij}$ is performed as:
\begin{equation}
    \Tilde{S}_{ij} = \begin{cases}
      1, & \text{if }  \Tilde{Y}_{ij} \geq \tau \\
      0, & \text{otherwise}
    \end{cases}
\end{equation}

This segment assignment is the final aggregated result of all the $K$ prompts. Unlike the best selection strategy, it does not suffer from the assumption that the highest confident result covers all the area from the ground truth.

\subsection{Post-processing}
\label{sectionPostProcessing}
To further refine the obtained segmentation mask, we  use a traditional computer vision morphology technique to enhance our results. By now, we have the aggregated result from the previous stage which were found to be composed of spatially close but potentially disjoint blobs. For this we use morphological closing \cite{gonzalez1992digital}. It helps us to connect disjoint blobs that are close to each other. 

\section{Results}
\label{sectionExperimentalResult}
We carried out experiments targeting two distinct classes: buildings and cars. These classes were selected due to the difficulty of detecting them in complex urban environments and their differing sizes. Therefore, by focusing on these two classes, we can comprehensively evaluate the effectiveness of our proposed segmentation method. 3457 test scenes/images per target class are sampled from the ISPRS Potsdam dataset \cite{rottensteiner2012isprs}. 
\par
We used Intersection over Union (IoU) as our metric to evaluate the performance of our model. For an image with size $(m \times n)$, let its ground truth matrix be $G$ and its predicted segmentation be denoted by $P$. For a class $c$, the IoU score is calculated as shown in Equation \ref{eqIou}.

\begin{equation}
\begin{split}
    \text{IoU}(G, P) = \sum_{i=1}^{m} \sum_{j=1}^{n} \frac{1\{G_{ij}=c \text{ and } P_{ij}=c\}}{1\{G_{ij}=c \text{ or } P_{ij}=c\}} \\
    \text{where, } 1\{\text{condition}\} = 
    \begin{cases}
      1, & \text{if condition is true} \\
      0, & \text{otherwise}
    \end{cases}
\end{split}
\label{eqIou}
\end{equation}

\subsection{Results for building}
\begin{figure}[t!]
\centering
\begin{subfigure}{0.7\linewidth}
\centering
         \fbox{\includegraphics[height=1.5 cm]{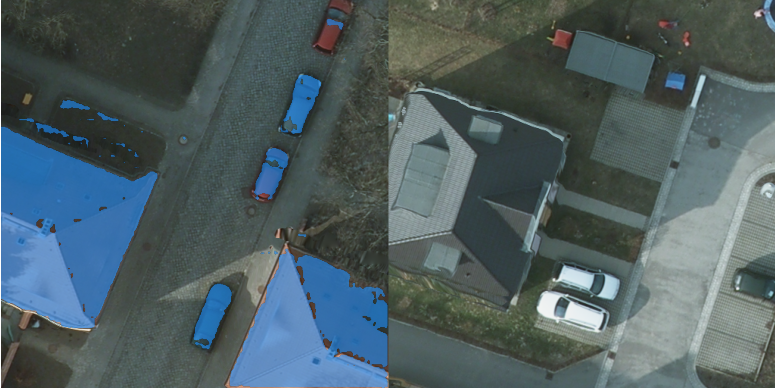}}
            \caption{}
            \label{figureResultBuildingPrompt1}
\end{subfigure}

\begin{subfigure}{0.7\linewidth}
\centering
         \fbox{\includegraphics[height=1.5 cm]{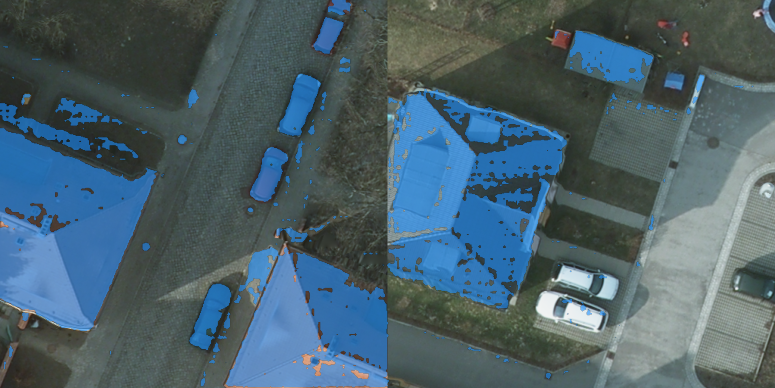}}
            \caption{}
            \label{figureResultBuildingPrompt2}
\end{subfigure}

\begin{subfigure}{0.7\linewidth}
\centering
         \fbox{\includegraphics[height=1.5 cm]{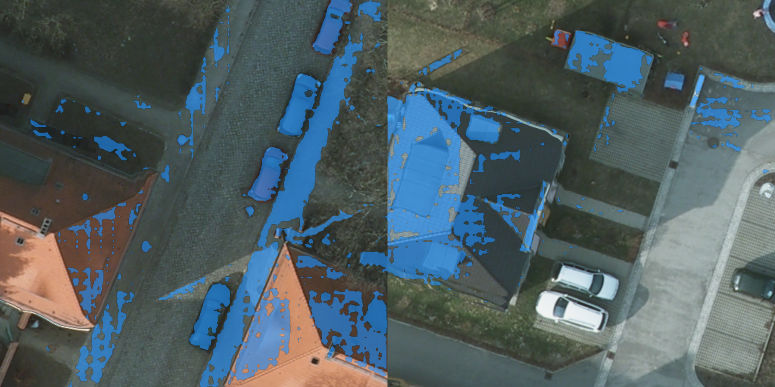}}
            \caption{}
            \label{figureResultBuildingPrompt3}
\end{subfigure}

\begin{subfigure}{0.7\linewidth}
\centering
         \fbox{\includegraphics[height=1.5 cm]{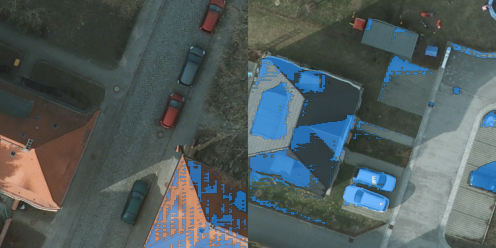}}
            \caption{}
            \label{figureResultBuildingPrompt4}
\end{subfigure}

\caption{Building detection on the stitched image (left: key image, right: query/test image) using four different proposed prompts shown in sub-figures (a), (b), (c) and (d)}
\label{figureBuildingDetectionResult}
\end{figure}

\begin{figure}[t!]
\centering
\begin{subfigure}{0.7\linewidth}
\centering
         \fbox{\includegraphics[height=1.5 cm]{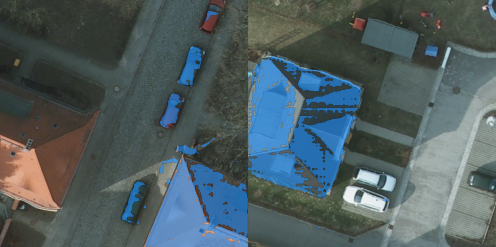}}
            \caption{}
            \label{figureConfidenceHighBuilding}
\end{subfigure}

\begin{subfigure}{0.7\linewidth}
\centering
         \fbox{\includegraphics[height=1.5 cm]{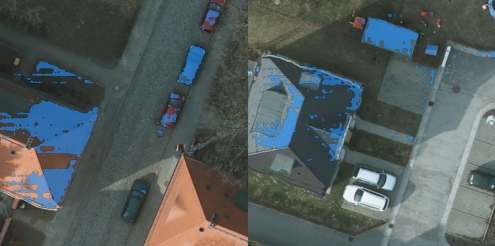}}
            \caption{}
            \label{figureConfidenceLowBuilding}
\end{subfigure}

\caption{Building segmentation masks for two different confidence scores: (a) High - 0.8, (b) Low - 0.4.}
\label{figureConfidenceScoreComparison}
\end{figure}

A result for the building class is visualized in Figure \ref{figureBuildingDetectionResult}. In the stitched image, left half represents the key image and the right half represents the test/query image. Essentially, we are interested to obtain good segmentation result on the right half only, as the result for the key image (left half) is known to us. However, the key-only prompt technique produces a spatially biased result that almost excludes the right half, as shown in Figure \ref{figureResultBuildingPrompt1}. This shows that even though the key only prompt technique is most intuitive, the result produced by it is not useful for our task. On the other hand, by introducing positive prompts from the query, building detection accuracy on the query image improves significantly (Figure \ref{figureResultBuildingPrompt2}). As discussed in Section \ref{sectionEnsemble}, the proposed method obtains segmentation masks $K$ times with different prompts and then obtains the final result by aggregating them. A comparison of two different segmentation masks with different scores are shown in Figure \ref{figureConfidenceScoreComparison}. This illustrates the relationship between better score and better segmentation performance on the test scene. The other proposed prompt technique (negative prompts from key and positive prompts from the test) performs similarly to the second prompt technique (Figure \ref{figureResultBuildingPrompt3}). However, for the proposed fourth prompt technique, the results degraded due to the imbalance between the key and query prompts (Figure \ref{figureResultBuildingPrompt4}). The fourth technique adds an input mask prompt to the second technique which one may expect intuitively to work better. But due to overindulgence of key prompts, i.e. introduction of key input mask with already present positive and negative key points created a spatial imbalance between key and query prompts leading to poorer results. The quantitative result is shown in Table \ref{tableResultBuilding}.
\par
We must note that with three advanced prompt engineering techniques, the performance on the key image (left part of the stitched image) actually degrades. However, this is not a concern to us, since the segmentation mask of the key image is known to us and we are not trying to find it here.
\par
We can also see the effect of CWMV strategy of aggregation. In Figure \ref{figureConfidenceHighBuilding}, we can see the result of a prompt with highest confidence score. Even with highest confidence, it still misses out on providing segment of the second smaller building next to it. That building is identified in a lower confident result, which can be seen in Figure \ref{figureConfidenceLowBuilding}, albeit with some false positives. The selection of the best result would have deprived us of detection of the smaller building, which is where CWMV strategy comes into the picture. CWMV aggregates the results of all the prompts and eliminates the pixels identified as building with low aggregated confidence score. The result of this can be seen in Figure \ref{figureResultBuildingPrompt2} where we can observe that both the buildings are well segmented and the false positives from Figure \ref{figureConfidenceLowBuilding} are also eliminated. This strategy clearly helped us aggregate the prompt results in a more efficient manner.
\par
In order to compare our method with a supervised method, we took a UNet based model with the backbone of ResNet-34 and pretrained on the ImageNet dataset. The model was fine-tuned on one labelled example; the same example that was used by our method as the key image. The fine-tuned model was then tested on the unseen test set which gave an IoU score of 0.3214. Thus, our SAM-based method proved to work better than a pretrained UNet.
\par
\begin{figure}[t!]
\centering
\begin{subfigure}{0.45\linewidth}
\centering
    \fbox{\includegraphics[height=1.5cm]{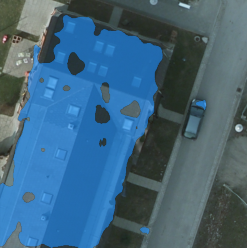}}
    \caption{}
    \label{figureResultRaw1}
\end{subfigure}
\begin{subfigure}{0.45\linewidth}
\centering
    \fbox{\includegraphics[height=1.5cm]{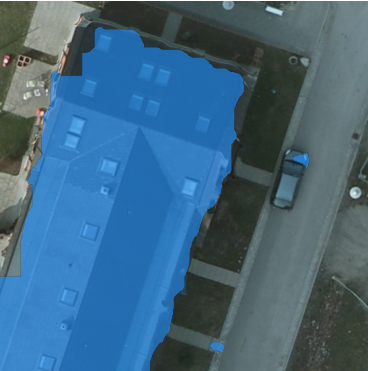}}
    \caption{}
    \label{figureResultProcessed1}
\end{subfigure}
\begin{subfigure}{0.45\linewidth}
\centering
    \fbox{\includegraphics[height=1.5cm]{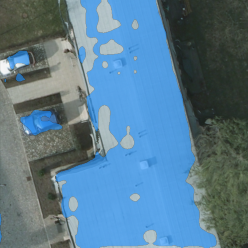}}
    \caption{}
    \label{figureResultRaw2}
\end{subfigure}
\begin{subfigure}{0.45\linewidth}
\centering
    \fbox{\includegraphics[height=1.5cm]{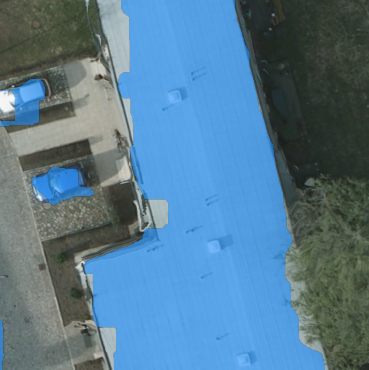}}
    \caption{}
    \label{figureResultProcessed2}
\end{subfigure}
\caption{Effect of the closing morphology. Figures (a) and (c) show the results after applying the CWMV aggregation strategy. Figures (b) and (d) show the result after passing them through the closing morphology as the blobs left behind by CWMV are filled up.}
\label{figurePostprocessing}
\end{figure}

We observe the the results of our post processing using  morphology in Figure \ref{figurePostprocessing}. The results of this can be clearly seen by comparing figures \ref{figureResultRaw1} and \ref{figureResultRaw2} (where the results from CWMV had blobs unidentified due to its pixelwise thresholding) with figures \ref{figureResultProcessed1} and \ref{figureResultProcessed2} (after passing through closing morphology). The morphology helped us provide more complete segments as results and increased the efficiency of our overall methodology.
\par
\subsection{Results for car}
\begin{figure}[t!]
\centering
\begin{subfigure}{0.7\linewidth}
\centering
         \fbox{\includegraphics[height=1.5 cm]{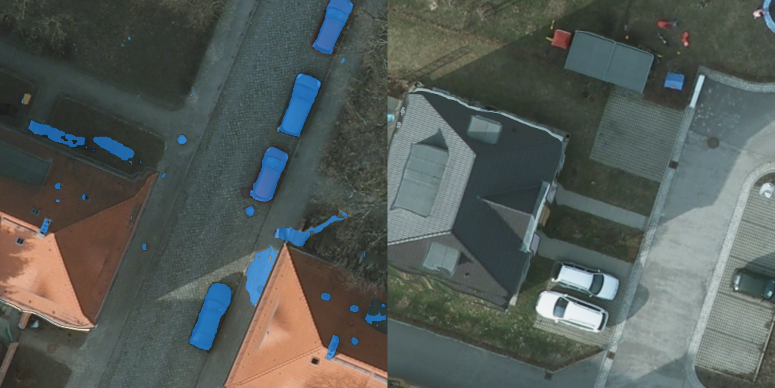}}
            \caption{}
\end{subfigure}

\begin{subfigure}{0.7\linewidth}
\centering
         \fbox{\includegraphics[height=1.5 cm]{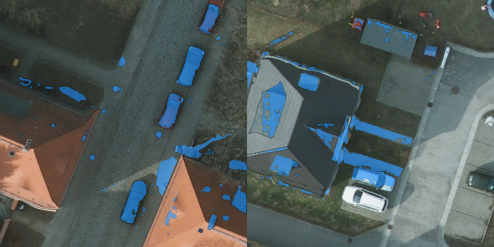}}
            \caption{}
\end{subfigure}

\begin{subfigure}{0.7\linewidth}
\centering
         \fbox{\includegraphics[height=1.5 cm]{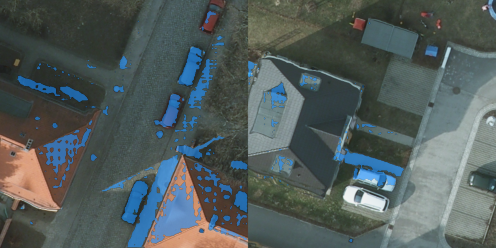}}
            \caption{}
\end{subfigure}

\begin{subfigure}{0.7\linewidth}
\centering
         \fbox{\includegraphics[height=1.5 cm]{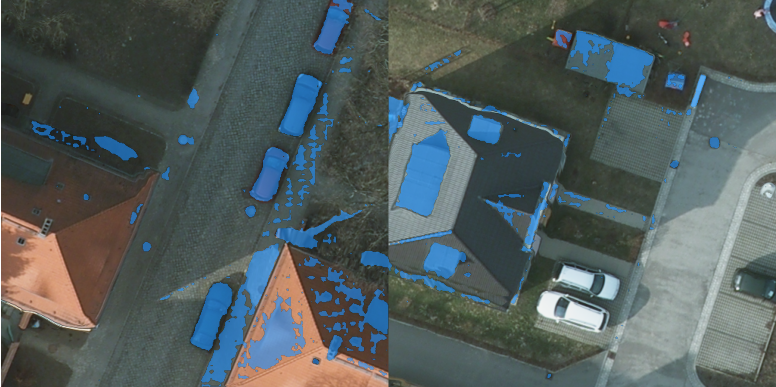}}
            \caption{}
\end{subfigure}

\caption{Car detection on the stitched image (left: key image, right: query/test image) using four different proposed prompts shown in sub-figures (a), (b), (c) and (d).}
\label{figureCarDetectionResult}
\end{figure}

A result for the car class is illustrated in Figure \ref{figureCarDetectionResult}. Similar to the previous observations, the key-only prompt technique yields a spatially biased result, almost excluding the right half. In contrast, the incorporating positive prompts from the target enhances car detection accuracy in the query image. The quantitative results are presented in Table \ref{tableResultCar}. Overall, the accuracy for car detection is significantly lower compared to building detection, which shows the greater challenge posed by the smaller size of cars in the analyzed scenes.

\begin{table}[!h]
    \centering
    \begin{tabular}{|c|c|c|c|c|}
        \hline
         Methods &  \multicolumn{2}{|c|}{Best Selection} & \multicolumn{2}{|c|}{CWMV} \\
         \cline{2-5}
         \; & Raw & Processed & Raw & Processed \\
         \hline
        Prompt 1 & 0.0008 & 0.0015 & 0.0011 & 0.0019\\
        Prompt 2 & 0.5662 & 0.6368 & \textbf{0.6485} & \textbf{0.6930}\\
        Prompt 3 & 0.2890 & 0.3323 & 0.2996 & 0.3954\\
        Prompt 4 & 0.3918 & 0.4690 & 0.4372 & 0.4796 \\
        \hline
    \end{tabular}
    \caption{Quantitative performance of proposed prompt engineering techniques for the building class, shown as Intersection over Union (IoU).}
\label{tableResultBuilding}
\end{table}

\begin{table}[!h]
    \centering
    \begin{tabular}{|c|c|c|c|c|}
        \hline
         Methods &  \multicolumn{2}{|c|}{Best Selection} & \multicolumn{2}{|c|}{CWMV} \\
         \cline{2-5}
         \; & Raw & Processed & Raw & Processed \\
         \hline
        Prompt 1 & 0.0000 & 0.0000 & 0.0000 & 0.0000\\
        Prompt 2 & 0.0067 & 0.0109 & \textbf{0.0237} & \textbf{0.0451}\\
        Prompt 3 & 0.0005 & 0.0008 & 0.0021 & 0.0031\\
        Prompt 4 & 0.0020 & 0.0033 & 0.0135 & 0.0151 \\
        \hline
    \end{tabular}
    \caption{Quantitative performance of proposed prompt engineering techniques for the car class, shown as Intersection over Union (IoU).}
\label{tableResultCar}
\end{table}

\section{Conclusions}
\label{sectionConclusion}
Foundation models are advancing significantly in computer vision due to their strong generalization capabilities. This paper demonstrated the generalizability of SAM, a foundation model for semantic segmentation, when applied to remote sensing images. We highlighted its application in segmenting two challenging classes: buildings and cars. While the result for building is somewhat satisfactory, result for car needs significant improvement. Since research in this area is still in its early stages, this paper represents a step towards understanding how to leverage robust foundation models in Earth observation. Future work will further explore segmentation for challenging classes like cars and explore segmentation for low-resolution Earth observation images, which may present greater challenges due to their substantial differences from typical computer vision images.

%
%
\bibliographystyle{splncs04}
\bibliography{egbib}
\end{document}